\documentclass[letterpaper, 10 pt, conference]{ieeeconf}  % Comment this line out if you need a4paper

\usepackage{xcolor}
\usepackage{amsmath}
\usepackage{graphicx}
\usepackage{subfig}
\usepackage{multirow}
\usepackage{rotating}
\usepackage{soul}

\IEEEoverridecommandlockouts  % This command is only needed if you want to use the \thanks command

\title{\LARGE \bf
Data Augmentation for Automated Adaptive Rodent Training
}

\author{Dibyendu Das$^{1}$, Alfredo Fontanini$^{2}$, Joshua F. Kogan$^{2}$, Haibin Ling$^{1}$, C.R. Ramakrishnan$^{1}$, I.V. Ramakrishnan$^{1}$% <-this % stops a space
\thanks{$^{1}$Dibyendu Das, Haibin Ling, I. V. Ramakrishnan, and C. R. Ramakrishnan are with Department of Computer Science, Stony Brook University,
        Stony Brook, NY 11794, USA}%
\thanks{$^{2}$Alfredo Fontanini and Joshua F. Kogan are with the Department of Neurobiology and Behavior, Stony Brook University,
        Stony Brook, NY 11794, USA}%
}

\newcommand{\myparagraph}[1]{\medskip\noindent\textbf{#1:}\ }
\newcommand{\comment}[1]{}

\newcommand{\indiv}{\ensuremath \mathcal{S}}
\newcommand{\session}[1]{\ensuremath S_{#1}}
\newcommand{\trial}[2]{\ensuremath T_{#1,#2}}
\newcommand{\specific}[2]{\ensuremath #1^{(#2)}}
\newcommand{\norm}[1]{\ensuremath |#1|}

\begin{document}

\maketitle
\thispagestyle{empty}
\pagestyle{empty} %plain

%%%%%%%%%%%%%%%%%%%%%%%%%%%%%%%%%%%%%%%%%%%%%%%%%%%%%%%%%%%%%%%%%%%%
\begin{abstract}

Fully optimized automation of behavioral training protocols for lab animals like rodents has long been a coveted goal for researchers. It is an otherwise labor-intensive and time-consuming process that demands close interaction between the animal and the researcher. In this work, we used a data-driven approach to optimize the way rodents are trained in labs.
In pursuit of our goal, we looked at data augmentation, a technique that scales well in data-poor environments. Using data augmentation, we built several artificial rodent models, which in turn would be used to build an efficient and automatic trainer. Then we developed a novel similarity metric based on the action probability distribution to measure the behavioral resemblance of our models to that of real rodents.

% \newline

% \indent \textit{Clinical relevance} — \textcolor{red}{potential clinical trials using rodents for drug testing for neurological problems}

\end{abstract}

%%%%%%%%%%%%%%%%%%%%%%%%%%%%%%%%%%%%%%%%%%%%%%%%%%%%%%%%%%%%%%%%%%%%%%%%%%%%%%
\section{Introduction}

\myparagraph{Background} In lab experiments, operant conditioning tasks with food and(or) water rewards are commonly used to train and test rodents in a wide variety of sensory, motor and cognitive tasks. Water rewards can be dispensed with temporal and quantitative precision and consumed rapidly with minimal body movement (particularly so in head-fixed rodents), which makes them ideal for automated behavioral training, testing, and electro-physiological and optical recording.

One of the main barriers for research on complex behaviors in rodents, lies in training the animals -- a process typically done under close supervision of researchers who frequently modify protocols and procedures on an animal-by-animal basis to improve learning rates and performance. This approach is labor-intensive and time-consuming, and makes the interaction between animal and researcher an integral part of the training process, possibly confounding comparisons of experimental outcomes across animals and labs~\cite{crabbe99}.

Ideally, behavioral training systems should be fully automatic, ready to scale up, blind in design, and flexible in changing paradigms. There has been a long history of designing automatic behavior-training systems consisting of monitoring and feedback controlling. In freely moving mice, automatic measurement has been implemented in characterizing visual performance \cite{de05, benkner13, kretschmer13}, evaluation of pain sensitivity \cite{kazdoba07, roughan09}, freezing behavior during fear conditioning \cite{kopec07, anagnostaras10}, home-cage phenotyping \cite{hubener12, balci13}, anxiety \cite{aarts15}, and social behavior \cite{ohayon13, weissbrod13}.

\myparagraph{Motivation} 
Automatic training systems with multiple cognitive behaviors requiring memory, attention and decision making have been developed over the last few years in free-moving \cite{erlich11, poddar13} and head-fixed \cite{han18} rodents. However, majority of these works \cite{han18, poddar13, bjaanes18} have been confined to automating the existing training protocols using software and/or hardware tools. Tailoring the training protocols to individual animals to shorten training time has not been considered before.  \emph{This paper takes a step towards generating tailored training protocols using data-driven techniques.}

In most existing training procedures, the rodents are trained using a (pseudo-)randomly generated sequence of training inputs (gustatory, olfactory etc.) over a period of several days or even weeks.
%Previously no attention has been paid in the generation of these input sequences. 
Our long-term goal is to make the training protocol adaptive, i.e. to algorithmically \emph{change the sequence of training inputs based on each rodent's current performance}, so that they can successfully complete the training  as early as possible.
This adaptive strategy will be derived based on historical training data of similar animals under similar conditions, by leveraging computational techniques such as Reinforcement Learning.

%\textcolor{red} {What is the overall approach to the solution.}
\myparagraph{Approach}  At a high level, we cast the problem of determining an individually-tailored sequence of training inputs as an instance of a \emph{sequential decision making problem}, as follows.   The training inputs are regarded as the sequence of actions taken by a ``training agent'', based on observing the ``environment'': in this case, a trainee rodent's past behavior.  The trainer agent gets a ``reward'' that depends on the length of input sequence needed to successfully train a rodent.  A rodent's behavior is conditioned on a number of variables, many of which (e.g. descriptions of perceptive and cognitive states) cannot be measured; hence a rodent's behavior itself is modeled as a stochastic process.  Thus the rodent training problem is posed as one of determining the sequence of actions of the training agent that maximizes the expected reward.  Reinforcement Learning (RL) is a widely-used technique for determining optimal action sequences of agents when the environment's state space is not known \emph{a priori}, but can only be sampled. 

RL has been used successfully in a variety of problems ranging from robotics~\cite{xie18} to operations research and games~\cite{mnih15}.  However, we face a significant hurdle when considering RL for the rodent training problem: data-driven techniques like RL require copious data --- either historical data in large data sets, or the ability to generate samples on demand.  Historical rodent training data is limited; data on even hundreds of individual rodents is woefully inadequate for RL-based techniques.  Insufficient data results in unstable or sub-optimal solutions.  The usual method for deploying RL and other machine learning techniques for data-poor problems is to perform ``data augmentation'': adding artificially creating data points based on a domain-specific model.  For instance, in computer vision tasks (e.g. object recognition from images),  transformations such as rotation, translation and cropping are used to generate more data from a given data set, improving the stability and performance of the learned task~\cite{shorten19}.  

\emph{This paper describes a data augmentation technique to enlarge rodent behavioral data.}  Such an augmented data set enables use of RL for the rodent training problem.  In particular, we  describe an \emph{artificial rodent model} whose behavior, when subjected to similar training protocols, resembles that of a real rodent.  We bring rigor to the notion of ``resemblance'' by quantifying the degree of behavioral similarity of our artificial rodent with respect to a biological one, using a novel similarity metric based on \textit{action probability distribution}.  A well-defined metric for similarity will enable us to revise or refine the artificial rodent model and help ensure that the solutions to the rodent training problem derived using the augmented data will perform well when applied to the training of real rodents. 

\comment{
\textcolor{red}{What is novel about this approach it, why should somebody care about this.}

The idea of adaptively generating training protocols for lab animals (rodents) is quite new in the literature. In our experiments, we came up with several artificial rodent models that had done a decent job in successfully mimicking the behaviour of biological rodents. To identify the best model among those, we designed a novel similarity metric based on action probability distribution, using which we were able to have a mechanism for reliably comparing the degree of similarity between model behaviors.
}

\section{Training Setup}

The rodents are kept water restricted in a dark environment and habituated to head-restraint. Training is divided across multiple sessions with each session performed on a different day. Each training session is composed of a number of trials. In initial training sessions, rodents are presented gustatory stimuli (100mM Sucrose, 100mM NaCl). Once animals are trained on this discrimination, mixtures are introduced and rodents must learn to discriminate 55\%/45\% from 45\%/55\% (Sucrose\%/NaCl\%). This training frequently takes longer than 30 sessions(days) to complete. At the start of each trial a stimulus is presented to the rodents via a central spout kept in front of their mouth. The spout retracts after giving them the chance to lick the stimulus from its tip. After 3-second delay two other lateral spouts advance from both left and right and are kept within the reach of the rodents.  Water reward is presented at the left or right spout based on the category of the presented stimulus.  Rodents licking the correct spout collect the reward and the process continues. Otherwise, for an incorrect decision, they are penalized by a longer wait interval without water, before the next trial resumes.

The session ends when the rodents stop interacting with the spout(s). Typically one session lasts for about 150 trials on average. The entire training procedure ends successfully when a rodent performs with an accuracy of 70\% or more for three consecutive sessions. %\blue{Typically, on average, the training continues for about 20-30 sessions (days)}\footnote{30 sessions in the blue line above was added by Josh. In our data-set we've got maximum 21 sessions} before a rodent is declared \emph{trained}.

The sequence of taste stimuli generated to train a rodent is (pseudo-)random with a constraint that no more than 3 consecutive inputs are identical.

To make the training process more efficient, we aim to optimize this long sequence of sessions, by reducing the number of sessions(days). \comment{In the process of achieving this goal, we realized that the number of sessions could be reduced significantly if, instead of proving the rodents with completely random inputs, we could provide them with algorithmically generated adaptive sequences of inputs.}

\comment{
Computational techniques like Artificial Intelligence (AI) can be leveraged to generate the training sequences adaptively. But most modern-day AI techniques require massive amount of historical data to work reliably. Unfortunately, in case of behavioral training, limitation of historical training data has been a major challenge for us.

To work around this limitation, we focused on \emph{data augmentation}, a technique that we used to come up with several artificial rodent models that could reliably mimic the behaviour of the real rodents. To identify the best model among them, we designed a novel similarity metric based on action probability distribution and rigorously quantified the degree of similarity in behavior of our artificial rodents with respect to the biological ones.
}

\section{Modelling the Behavior of Rodents using Reinforcement Learning}

\begin{table}[ht]
\caption{Structure of one training sequence}
\label{training data}
\begin{center}
\begin{tabular}{cc|c|c|c|c|}
& & \multicolumn{4}{c|}{Trial \#} \\
\hline
\multirow{6}{1em}[-1em]{\begin{sideways}Session \#\end{sideways}} & 1 & $T_{1,1}$ & $T_{1,2}$ & $\cdots$ & $T_{1,m_{_1}}$ \\
\cline{2-6} & 2 & $T_{2,1}$ & $T_{2,2}$ & $\cdots$ & $T_{2,m_{_2}}$ \\
\cline{2-6} & & \multicolumn{4}{c|}{$\vdots$} \\
\cline{2-6} & j & $T_{j,1}$ & $T_{j,2}$ & $\cdots$ & $T_{j,m_j}$ \\
\cline{2-6} & & \multicolumn{4}{c|}{$\vdots$} \\
\cline{2-6} & n & $T_{n,1}$ & $T_{n,2}$ & $\cdots$ & $T_{n,m_n}$ \\
\hline
\end{tabular}
\end{center}
\end{table}

In our modelling, we used the following symbols for the gustatory stimuli:
\begin{align*}
    \mathrm{sweet}&\doteq100\mathrm{mM\ Sucrose}\\
    \mathrm{salt}&\doteq100\mathrm{mM\ NaCl}\\
    \mathrm{sweet\_55\%}&\doteq55\%/45\%\mathrm{\ (Sucrose\%/NaCl\%)}\\
    \mathrm{salt\_55\%}&\doteq45\%/55\%\mathrm{\ (Sucrose\%/NaCl\%)}
\end{align*}

The training data for a rodent is the temporal sequence of $n$ consecutive training sessions $\indiv = \langle \session{1}, \session{2}, \ldots, \session{n}\rangle$, where the $j$-th session, $\session{j}$ is a temporal sequence of $m_j$ consecutive trials, i.e. $\session{j}=\langle \trial{j}{1}, \trial{j}{2}, \ldots, \trial{j}{m_j}\rangle$ and the $k$-{th} trial $\trial{j}{k}$ is a triple $(\sigma_{j,k}, \rho_{j,k}, \lambda_{j,k})$ with stimulus $\sigma_{j,k} \in \{\mathrm{sweet}, \mathrm{sweet\_55\%}, \mathrm{salt\_55\%}, \mathrm{salt}\}$, response $\rho_{j,k} \in \{\mathrm{left}, \mathrm{right}, \mathrm{none}\}$, and outcome $\lambda_{j,k} \in \{\mathrm{correct}, \mathrm{incorrect}\}$.  We use superscripts (such as $\specific{\session{j}}{a}, \specific{\trial{j}{k}}{a},$ etc.) to denote data for individual rodent ``$a$'' in order to distinguish between data for different rodents.
The structure of the training data is shown in Table~\ref{training data}.
%\footnote{$n$ typically varies in the range between 12 to 21, with 12 being the highest session for which we've had training data for all the rodents. For each session $j$, $m_j$ typically varies in the range between 32 to 245, with mean, mode, and median being 153.63, 200, and 158 respectively.}
%\magenta{We've had 7 such training data sets $\{\specific{\indiv}{1}, \specific{\indiv}{2}, \ldots, \specific{\indiv}{7}\}$ to work with.}\footnote{We initially had 7, but after excluding 2 outliers, we were left with 5. Do you want me to change it to 5?}

\begin{figure*}
    \centering
    \includegraphics[width=\textwidth,keepaspectratio]{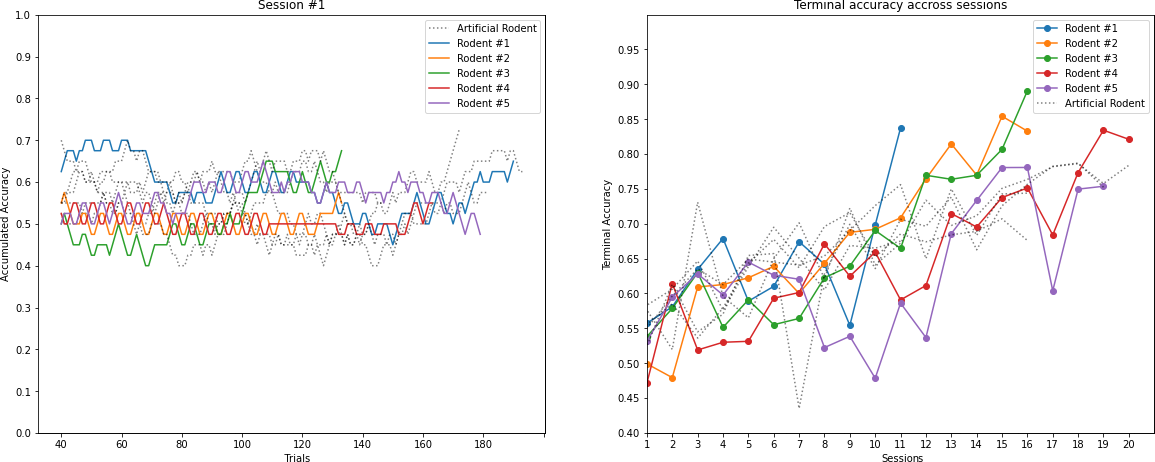}
    \caption{Left: sliding window mean 
    accuracy of the artificial and the real rodents across all the trials in session 1 (Black dots represent the model). Right: Terminal accuracy of the artificial and the real rodents in all the sessions.}
    \label{accuracy graph for simulation}
\end{figure*}

% \begin{figure}
%     \centering
%     \subfloat[\centering Session 1]{{\includegraphics[width=0.5\textwidth,keepaspectratio]{images/simulation_1.png} }}
%     \vspace{0.48cm}
%     \subfloat[\centering Session 6]{{\includegraphics[width=0.48\textwidth,keepaspectratio]{images/simulation_6.png} }}
%     \vspace{0.5cm}
%     \subfloat[\centering Session 12]{{\includegraphics[width=0.48\textwidth,keepaspectratio]{images/simulation_12.png} }}
%     \caption{Moving-mean accuracy of real and artificial model-rodent across all the trials in three different sessions: 1, 6 and 12 (Black dots represent the artificial model-rodent)}
%     \label{accuracy graph for simulation}
% \end{figure}

% \begin{figure}
%     \centering
%     \includegraphics[width=0.5\textwidth,keepaspectratio]{images/terminal_accuracy.png}
%     \caption{Moving mean accuracy over the last five trials for real and artificial rodents across all the session (Black dots represent the artificial rodent)}
%     \label{accuracy graph for simulation}
% \end{figure}

\comment{To work around this limitation, we focused on data augmentation, a technique that we used to come up with several artificial rodent models that could reliably mimic the behaviour of the real rodents.

In the next section we've discussed the artificial rodent model that \emph{reliably} mimics the behavior of a real rodent.
}

\subsection{Artificial Rodent Model} \label{implementation}

As the entire training process is structured around the episodic training sequences, we model the artificial rodent as an RL agent, more specifically, based on classical Q-Learning\cite{watkins89}. Any RL agent can be modelled using a Markov Decision Process (MDP) with three of its components: \emph{states $\mathcal{S}^+$}, \emph{actions $\mathcal{A}$}, and \emph{rewards $\mathcal{R}$} defined as follows.

\subsubsection{States $\mathcal{S}^+$}

For our model rodent, we've defined the state of the environment (observation) as the tuple of last $k$ input stimuli. The set of states, $\mathcal{S}^+ = \{(\sigma_{t-k+1}, \sigma_{t-k+2},\cdots, \sigma_t)\ |\ t\ge k-1\}$ where each element in the tuple is a stimulus, i.e., one of $\{\mathrm{sweet}, \mathrm{sweet\_55\%}, \mathrm{salt\_55\%}, \mathrm{salt}\}$. We've used $k=3$ in our experiments. As the state space is discrete (with $4^k$ possible states), we can store the action-value function $q_\pi(s,a)$ in a table known as the \emph{Q-Table}, where $q_\pi(s,a)$ denotes the expected cumulative reward that the rodent model can get from state $s$ by taking action $a$ and following policy $\pi$ thereafter.

\subsubsection{Actions $\mathcal{A}$}

The set of all possible actions is $\mathcal{A}=\{\mathrm{left}, \mathrm{right}, \mathrm{none}\}$.

The model picks its actions using the $\epsilon$-greedy policy with respect to its learned Q-Table: with probability $\epsilon$ it chooses an action at random (explore); and with probability $1-\epsilon$ it chooses an action according to the Q-Table (exploit). To mimic the behaviour closely, the model starts off with high values of $\epsilon$ and gradually decreases it by a small term after every episode. An episode ends when the session ends.

It has been shown\cite{singh00} that a learning policy always converges to an optimal policy if it satisfies the GLIE (Greedy in the Limit with Infinite Exploration) conditions:
\begin{enumerate}
    \item If a state is visited infinitely often and each action in that state is chosen infinitely often.
    \item There is a small (but non-zero) exploration probability $\epsilon_j$ at the start of episode $j$.
    \item In the limit the learning policy is greedy (with probability 1) with respect to the learned Q-function, i.e. $\displaystyle\lim_{j\to\infty}\epsilon_j=0$
\end{enumerate}

In our implementation, we've used the following equation to set the exploration probability $\epsilon_j$ at the start of session (episode) $j$ $$\epsilon_j = \epsilon_s + e^{-0.025j} - 1$$ where $\epsilon_s=0.8$ is the starting exploration probability.

During every trial of any arbitrary session $j$, the model explores with probability $\epsilon_j$, i.e. it selects any one of the available actions from $\mathcal{A}$ with uniform probability. And with probability $1-\epsilon_j$ it exploits action $a\sim P(a;s)$ i.e. samples an action $a$ from the discrete probability distribution $$P(a;s)=\frac{e^{q_\pi(s,a)}}{\displaystyle\sum_{a'\in\mathcal{A}}e^{q_\pi(s,a')}}$$ where $s$ is the state of the environment at the start of the trial.

\begin{figure*}[t!]
    \centering
    \subfloat[Model Rodent\label{correlation:model}]{{\includegraphics[width=0.44\textwidth,keepaspectratio]{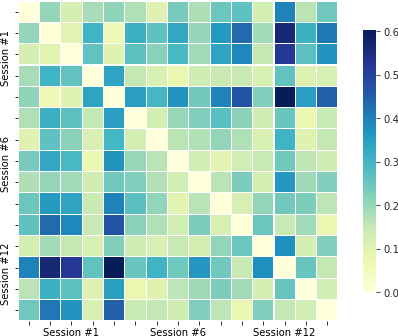} }}
    \hspace{2em}
    \subfloat[Real Rodents\label{correlation:real}]{{\includegraphics[width=0.44\textwidth,keepaspectratio]{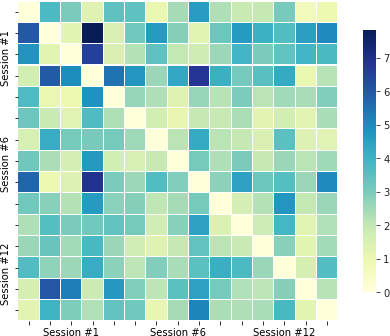} }}
    \caption{(a) Correlation heat-map of the similarity values (the lower the more similar) across multiple executions of the rodent model in each session. (b) Correlation heat-map of the similarity values between trajectories (sequence of trials) in each session. Data is obtained from five real rodents.}
\end{figure*}

\subsubsection{Reward Model $\mathcal{R}$}

For every correct and incorrect action the model is rewarded with +1 and -1 respectively. Just like what happens with the biological rodents, we stop the training when the model reaches a pre-defined accuracy of 70\% or more for three consecutive sessions.

At any particular trial $t$, let's say that the state of the environment is $s_t$, the agent takes an action $a_t$, receives the reward $r_t$ and thereby changes the state of the environment to $s_{t+1}$ in the next trial $t+1$. The action value function is then updated by the following Q-Learning update rule~\cite{watkins89}  $$q(s_t,a_t) = q(s_t,a_t) + \alpha\cdot[r_t + \gamma\cdot\max_{a'\in\mathcal{A}}\{q(s_{t+1},a')\} - q(s_t,a_t)]$$ Throughout our experiment, we used the learning rate $\alpha=0.2$ and the discount factor $\gamma=1.0$ (no discount).

\subsection{Similarity Metrics}
\subsubsection{Similarity between individual behaviors}
Let $a$ and $b$ be two individuals (data from real rodents, or two distinct executions of our above-described model rodent), and let $\specific{\session{i}}{a}$ and $\specific{\session{j}}{b}$ be data from the $i$-th session for individual $a$ and the $j$-th session from individual $b$, respectively.  We now define a metric to quantify the similarity between sessions $\specific{\session{i}}{a}$ and $\specific{\session{i}}{b}$.  This quantity is defined using a distance metric $\mathcal{D}$ over discrete probability distributions.  At a high level, we consider sliding windows of fixed length (denoted by $\Delta$) at various points over $\specific{\session{i}}{a}$ (and $\specific{\session{j}}{b}$),   the frequencies of correct actions as distributions, and define the similarity metric as the mean distance between such distributions.   Formally, the \emph{mean distance between windowed distribution of actions}, denoted by $E_{i,j}(a,b)$ is defined as:
\comment{
We simulated the model rodent on the same input sequences as used with the real rodents. For every session, the model was executed 5 times with identical initial conditions. We randomly selected two such executions from two arbitrary sessions to define our similarity metric. For any two randomly chosen executions $t_1$ and $t_2$ (out of the 10 executions of the model) from two particular sessions $i$ and $j$, let $S^{(t_1)}_i$ and $S^{(t_2)}_j$ be the sequences of trials, where $S^{(x)}_y = \left\langle T^{(x)}_{y,1}, T^{(x)}_{y,2}, \cdots, T^{(x)}_{y,{m_y}}\right\rangle$ and the $k^{th}$ trial $T^{(x)}_{y,k}$ is a tuple $\left(i^{(x)}_{y,k}, o^{(x)}_{y,k}, l^{(x)}_{y,k}\right)$, with $l^{(x)}_{y,k} \in \{\mathrm{correct}, \mathrm{incorrect}\}\quad\forall x\in\{t_1,t_2\}$ and $y\in\{i,j\}$. That is, $S^{(t_1)}_i$ and $S^{(t_2)}_j$ have $m_i$ and $m_j$ number of trials respectively. We define the error between $S^{(t_1)}_i$ and $S^{(t_2)}_j$ as a mean of windowed distribution of actions (normalized action histograms over a sliding window over trials) defined as follows
}
\comment{
\begin{multline}\label{similary}
E_{i,j}(a,b) = \frac{1}{L-\Delta+1}\times \\
    \sum_{t=0}^{L-\Delta}\mathcal{D}\left(\frac{1}{\Delta}\displaystyle\sum^\Delta_{l=1}\mathbf{1}_{[\sigma^{(a)}_{i,t+l} = \mathrm{correct}]}, \frac{1}{\Delta}\displaystyle\sum^\Delta_{l=1}\mathbf{1}_{[\sigma^{(b)}_{j,t+l} = \mathrm{correct}]}\right)
\end{multline}
}
\begin{align}\label{similary}
E_{i,j}(a,b) = \frac{1}{L-\Delta+1}\times 
    \sum_{t=0}^{L-\Delta}\mathcal{D}\left(\delta^{(a)}_{i,t}, \delta^{(b)}_{j,t}\right)
\end{align}
where
\begin{align}\label{windowed distribution}
\delta^{(\alpha)}_{i,t} = \frac{1}{\Delta}\displaystyle\sum^\Delta_{l=1}\mathbf{1}_{[\lambda^{(\alpha)}_{i,t+l} = \mathrm{correct}]}
\end{align}
%
% \begin{multline}\label{similary}
% E^{(t_1,t_2)}_{i,j} = \frac{1}{\min(m_i,m_j)-\Delta+1}\times \\
%     \sum_{t=0}^{\min(m_i,m_j)-\Delta}\mathcal{D}\left(\left(\cdot\displaystyle\sum^\Delta_{l=1}\mathbf{1}_{[o^{(t_1)}_{i,t+l} = \mathrm{left}]},\frac{1}{\Delta}\cdot\displaystyle\sum^\Delta_{l=1}\mathbf{1}_{[o^{(t_1)}_{i,t+l} = \mathrm{right}]}\right)\right.,\\
%     \left.\left(\frac{1}{\Delta}\cdot\displaystyle\sum^\Delta_{l=1}\mathbf{1}_{[o^{(t_2)}_{j,t+l} = \mathrm{left}]},\frac{1}{\Delta}\cdot\displaystyle\sum^\Delta_{l=1}\mathbf{1}_{[o^{(t_2)}_{j,t+l} = \mathrm{right}]}\right)\right)
% \end{multline}
%
where $\mathcal{D}$ is the chosen distance function over distributions,  $L=\min\left(\norm{\specific{\session{i}}{a}}, \norm{\specific{\session{j}}{b}}\right)$, $\Delta$ is the window length and $\mathbf{1_{[x=y]}}$ is the indicator function defined as:
\[
    \mathbf{1_{[x=y]}} = \left\{
        \begin{array}{ll}
            1 & \text{if } x=y\\
            0 & \text{otherwise}
        \end{array}\right.
\]

\noindent
 Any distance metric can be used to measure the distance between two discrete probability distributions. In our results we've used \emph{Match Distance} ($L_1$ distance) defined as follows:

\begin{align*}
    \mathcal{D}\left(D^{(1)}, D^{(2)}\right) &= \sum_{i \in \mathit{dom}}\left|\mathit{pmf}^{(1)}(i) - \mathit{pmf}^{(2)}(i)\right|
\end{align*}

\noindent
where the two distributions $D^(1)$ and $D^(2)$ have the same domain $\mathit{dom}$, and $\mathit{pmf}^{(1)}$ and $\mathit{pmf}^{(2)}$ are the probability mass functions of the two distributions.

\subsubsection{Similarity between group behaviors}
The inherently stochastic nature of the behaviors makes comparison of individuals challenging.  To identify trends more clearly, we can compare groups as a whole.  We define a comparison metric between group behaviors closely following the mean windowed distribution of actions, as follows.  Given a group of $N$ individuals $\{\alpha_1, \alpha_2, \ldots, \alpha_n\}$, we first define a \emph{sliding-window mean} of their $j$-th session as:
\[
    \hat{S}_j=\left\langle\hat{\delta}_{j,0},\hat{\delta}_{j,1},\ldots,\hat{\delta}_{j,{m_j}-\Delta}\right\rangle
\]
where 
\begin{align}
    \hat{\delta}_{j,t}=\frac{1}{N\Delta}\sum^N_{i=1}\sum^\Delta_{l=0}\mathbf{1}_{[\lambda^{(\alpha_i)}_{j,t+l} = \mathrm{correct}]}
\end{align}

\noindent We define the average distance between the group's behavior between sessions $i$ and $j$ as
\begin{align}\label{session similary}
E_{i,j} = \frac{1}{L-\Delta+1}\times \displaystyle\sum_{t=0}^{L-\Delta}\mathcal{D}\left(\hat{\delta}_{i,t}, \hat{\delta}_{j,t}\right)
\end{align}
 where $L=\min\left(\norm{\hat{S}_i}, \norm{\hat{S}_j}\right)$.

\comment{
For each execution $t_i$ (out of 20 executions of the model) in any particular session $j$, let $S^{(t_i)}_j$ be the sequences of $m_j$ trials, i.e. $S^{(t_i)}_j = \left\langle T^{(t_i)}_{j,1}, T^{(t_i)}_{j,2}, \cdots, T^{(t_i)}_{j,m_j}\right\rangle$ and the $k^{th}$ trial $T^{(t_i)}_{j,k}$ is a tuple $\left(i^{(t_i)}_{j,k}, o^{(t_i)}_{j,k}, l^{(t_i)}_{j,k}\right)$, with $l^{(t_i)}_{j,k}\in\{\mathrm{correct}, \mathrm{incorrect}\}$. We compute the set $\hat{S}_j$ of \emph{mean of sliding-window mean} of distributions as follow:
\[
    \hat{S}_j=\left\langle\hat{S}_{j,1},\hat{S}_{j,2},\cdots,\hat{S}_{j,n},\cdots,\hat{S}_{j,{m_j}-\Delta+1}\right\rangle
\]
}

\section{Experimental Results}

We ran our model on the same input sequence that were used to train the real rodents in the lab. Using black dots in  the Fig.~\ref{accuracy graph for simulation} we show the accuracy across all the trials for our model-rodent in three particular sessions -- 1, 6, and 12. We've also shown (in colored lines) the sliding-window moving-mean accuracy across all the trials for every real rodent in the same three particular sessions -- 1, 6, and 12.

Although, by visual inspection these accuracy curves of artificial rodents look very similar to that of the real ones, we still need to define the degree of similarity formally and compare the same with respect to the real rodents.

To identify the best model among them, we designed a novel similarity metric based on action probability distribution and rigorously quantified the degree of similarity in the behavior of our artificial rodents with respect to biological ones.

%\subsection{Correlation Matrix of Similarity}

We ran the artificial rodent-model for 12 consecutive sessions, 5 executions in each session with identical initial conditions. Then we compared the similarity between each such execution using the previously defined Equation~\ref{similary}. Fig.~\ref{correlation:model} shows the correlation matrix of similarity values (higher values indicate lower similarity) for the rodent model, between any two executions from three arbitrarily chosen sessions -- session 1, 6, and 12. In Fig.~\ref{correlation:real} we show the corresponding values for five real rodents computed in a similar way.

% We also compared the performance of three rodent models by varying their initial exploration probabilities $\epsilon_s$ (the significant meta parameter of our model) and ran them 10 times in each session up to a total of 12 sessions. These initial exploration probabilities were set to 0.35, 0.65, and 1.0 for model \#1, model \#2, and model \#3 respectively. Fig.~\ref{model correlation} shows the correlation matrix of errors in the last session (session \#12) for these three models.

\begin{figure}
    \centering
    \includegraphics[width=0.46\textwidth,keepaspectratio]{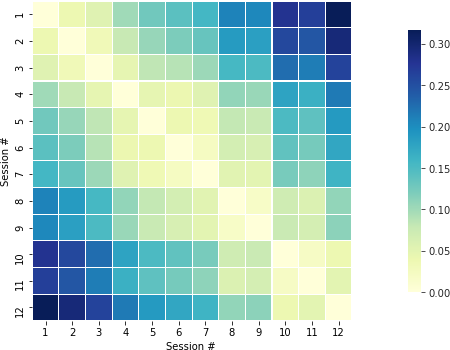}
    \caption{Correlation heat-map of the similarity values(lower values are more similar) between multiple sessions (average over each sliding window) for the rodent model.}
    \label{average session correlation}
\end{figure}

Fig.~\ref{average session correlation} shows group behavior of our model rodents compared across multiple sessions.
The figure shows the average distance between each pair of sessions using equation~\ref{session similary}, where each cell $(i,j)$ represents the average distance in sessions $i$ and $j$. The figure shows a clear trend in session-wise behavior as we increase the number of samples (we used 20 samples per session to generate the plot).

\section{Conclusions}

Behavioral training of lab animals has always been an arduous task for researchers. Optimization of these training protocols while minimizing their dependencies on human trainers, has been the major bottle-neck. With the recent advancement of Artificial Intelligence techniques, our data driven approach towards accomplishing this goal will create new paths for researchers to explore. Usage of data augmentation to built artificial rodent models will allow us to use these models in future to build an adaptive and intelligent trainer agent that will reduce the human intervention to a great extent.

\addtolength{\textheight}{-12cm}   % This command serves to balance the column lengths on the last page of the document manually. It shortens the text-height of the last page by a suitable amount. This command does not take effect until the next page so it should come on the page before the last. Make sure that you do not shorten the text-height too much.

% \begin{figure}
%     \centering
%     \includegraphics[width=0.42\textwidth,keepaspectratio]{images/3.png}
%     \caption{Correlation heat-map  of  the  similarity values (lower values are more similar) between multiple sessions (average over each sliding window) for the five real rodents.}
% \end{figure}

% \begin{figure}
%     \centering
%     \includegraphics[width=0.42\textwidth,keepaspectratio]{images/4.png}
%     \caption{Correlation heat-map  of  the  similarity values between multiple sessions (average over each sliding window) for five real rodents (x-axis) and five sample trajectories from the model rodent (y-axis).}
% \end{figure}

%%%%%%%%%%%%%%%%%%%%%%%%%%%%%%%%%%%%%%%%%%%%%%%%%%%%%%%%%%%%%%%%%%%%%%%%%%%%%%%%

\bibliographystyle{unsrt}
\bibliography{main}

\end{document}